\documentclass[conference]{IEEEtran}

\usepackage{graphicx}
\usepackage[usenames,dvipsnames]{xcolor}
\usepackage{url}  
\usepackage{psfrag}  

\begin{document}

\IEEEoverridecommandlockouts
\IEEEpubid{\makebox[\columnwidth]{978-1-4799-8622-4/15/\$31 \copyright 2015 IEEE \hfill} \hspace{\columnsep}\makebox[\columnwidth]{ }}

\title{Evaluating Go Game Records for Prediction of Player Attributes}

\author{\IEEEauthorblockN{Josef~Moud\v{r}\'{i}k}
\IEEEauthorblockA{ Charles University in Prague\\ Faculty of Mathematics and Physics\\
Malostransk\'e n\'am\v{e}st\'\i~25\\
Prague, Czech Republic\\
email: j.moudrik@gmail.com}%
\and\IEEEauthorblockN{Petr~Baudi\v{s}}
\IEEEauthorblockA{Czech Technical University\\ Faculty of Electrical Engineering\\
Karlovo n\'am\v{e}st\'\i~13\\
Prague, Czech Republic\\
email: pasky@ucw.cz}%
\and\IEEEauthorblockN{Roman Neruda}%
\IEEEauthorblockA{Institute of Computer Science\\ Academy of Sciences of the Czech Republic\\
Pod Vod\'{a}renskou v\v{e}\v{z}\'\i~2\\
 Prague, Czech Republic\\
email: roman@cs.cas.cz}
}

\maketitle

\begin{abstract}
We propose a~way of extracting and aggregating per-move evaluations
from sets of Go game records. The evaluations capture different aspects
of the games such as played patterns or statistic of sente/gote
sequences. Using machine learning algorithms, the evaluations can
be utilized to predict different relevant target variables. We apply this
methodology to predict the strength and playing style of the player
(e.g. territoriality or aggressivity) with good accuracy.
%
We propose a number of possible applications including aiding in Go study, 
seeding real-work ranks of internet players or tuning of Go-playing programs.
\end{abstract}


\section{Introduction}
The field of computer Go is primarily focused on the problem
of creating a~program to play the game by finding the best move from a~given
board position~\cite{GellySilver2008}. We focus on analyzing existing game
records with the aim of helping humans to play and understand the game better
instead.

Go is a~two-player full-information board game played
on a~square grid (usually $19\times19$ lines) with black and white
stones; the goal of the game is to surround territory and
capture enemy stones. In the following text we assume basic
familiarity with the game rules, a~small glossary can be found
at the end of the paper.

Following up on our initial research~\cite{GoStyleArxiv},
we present a method for extracting information from game records.
We extract different pieces of domain-specific
information from the records to create a complex
evaluation of the game sample. The \emph{evaluation} is a vector
composed of independent features -- each of the features
captures different aspect of the sample. For example,
a statistic of most frequent local patterns played, or a statistic of high and
low plays in different game stages are used.

Using machine learning methods, the evaluation of the sample
can be used to predict relevant variables. In this work
in particular, the data sample consists of games of a player, and it is 
used to predict the player's strength and playing style.

This paper is organized as follows. Section~\ref{sec:rel}
summarizes related work in the area of machine learning applications 
in the game of Go. Section~\ref{sec:feat} presents the features comprising the evaluation.
Section~\ref{sec:mach} gives details about the machine
learning method we have used.
In Section~\ref{sec:expe} we give details about our
datasets -- for prediction of strength and style -- and
show how precisely can the prediction be conducted.
Section~\ref{sec:disc} discusses applications and future work.

%

\section{Related Work}
\label{sec:rel}
Since the game of Go has a worldwide popularity, large collections
of Go game records have been compiled, covering both amateur
and professional games, e.g.~\cite{KGS,GoGoD}.


Currently, the ways to utilize these records could be
divided into two directions. Firstly, there is the field of computer Go,
where the records have been used to rank certain patterns which serve as a
heuristic to speed up the tree-search~\cite{PatElo}, or to 
generate databases of standard openings~\cite{audouard2009grid}. They have also
been used as a source of training data by various neural-network
based move-predictors. Until very recently, these did not perform 
convincingly~\cite{enzenberger1996integration}. The recent
improvements~\cite{sutskever2008mimicking, clark2014teaching}
based on deep convolutional neural networks seem to be changing the situation and 
promising big changes in the field.

Secondly, the records of professional games have 
traditionally served as a study material for human players. 
There exist software tools~\cite{Kombilo,MoyoGo} designed to enable
the user to search the games. These tools also give statistics of next
moves and appropriate win rate among professional games.

Our approach seems to reside on the boundary between the two
above mentioned directions, with possible applications in both
computer Go and tools aiding in human study.
To our knowledge, the only work somewhat resembling ours is~\cite{CompAwar},
where the authors claim to be able to classify
player's strength into 3 predefined classes (casual,
intermediate, advanced player). In their work, the 
domain-specific features were extracted
by using GnuGo's~\cite{GnuGo} positional assessment
and learned using random forests~\cite{breiman01}.
It is hard to say how precise their method is, since
neither precision, nor recall was given.
The only account given were two examples of
development of skill of two players (picked in an unspecified
manner) in time.

One of the applications of hereby proposed methodology 
is a utilization of predicted styles of a player to recommend relevant
professional players to review. The playing style is
traditionally of great importance to human players, but
so far, the methodology for deciding player's style has been
limited to expert judgement and hand-constructed questionnaires
\cite{style:baduk}, \cite{senseis:which_pro}. See the Discussion
(Section~\ref{sec:disc}) for details.


\section{Feature Extraction}
\label{sec:feat}
This section presents the methods for extracting the evaluation 
vector (denoted $ev$) from a set of games. Because we aggregate
data by player, each game in the set is accompanied by the color
which specifies our player of interest.
The sample is therefore regarded as a \emph{set
of colored games}, $GC = \{ (game_1, color_1), \ldots\}$.

The evaluation vector $ev$ is composed by concatenating several
sub-vectors of \emph{features} -- examples include the
aforementioned local patterns or statistic of sente and gote
sequences. These will be described in detail in the rest of this section.
Some of the details are omitted, see~\cite{Moudrik13} for
an extended description.

\subsection{Raw Game Processing}
The games are processed by the Pachi Go
Engine~\cite{Pachi} which exports a variety of analytical data
about each move in the game.
For each move,
Pachi outputs a list of key-value pairs regarding the current move:

\begin{itemize}
    \item \emph{atari flag} --- whether the move put enemy stones in atari,
    \item \emph{atari escape flag} --- whether the move saved own stones from atari,
    \item \emph{capture} --- number of enemy stones the move captured,
    \item \emph{contiguity to last move} --- the gridcular
        distance (cf. equation~\ref{eq:gridcular}) from the last move,
    \item \emph{board edge distance} --- the distance from
        the nearest edge of the board,
    \item \emph{spatial pattern} --- configuration of stones around the played move.
\end{itemize}

We use this information to compute the higher level features given below.
The spatial pattern is comprised of positions of stones around
the current move up to a certain distance, given by the {\em gridcular} metric
\begin{equation}
d(x,y) = |\delta x| + |\delta y| + \max(|\delta x|, |\delta y|).
\label{eq:gridcular}
\end{equation}

This metric produces a circle-like structure on the Go board square
grid~\cite{SpatPat}. Spatial patterns of sizes 2 to 6 are taken into account.

\subsection{Patterns}
\label{feat:pat}
The pattern feature family is essentially a statistic of $N$ the most frequently occurring
spatial patterns (together with both atari flags). The list of the $N$ most frequently
played patterns is computed beforehand from the whole database of games. The patterns
are normalized so that it is black's turn, and they are invariant under rotation and mirroring.
We used $N=1000$ for the domain of strength and $N = 600$ for the domain of style
(which has a smaller dataset, see Section~\ref{res:style} for details).

Given a set of colored games $GC$ we then count how many times was each of the $N$
patterns played -- thus obtaining a vector $\vec c$ of counts ($|\vec c| = N$).
With simple occurrence count however, particular counts $c_i$ increase proportionally to 
number of games in $GC$. To maintain invariance under the number of games in the sample,
a normalization is needed. We do this by dividing the $\vec c$ by $|GC|$, though other
normalization schemes are possible, see~\cite{Moudrik13}.

\subsection{$\omega$-local Sente and Gote Sequences}
\label{feat:sente}
The concept of sente and gote is traditionally very important for human players, which
means it could bear some interesting information.
Based on this intuition, we have devised a statistic which tries to capture distribution of
sente and gote plays in the games
from the sample. In general, deciding what moves are sente or gote is hard. Therefore,
we restrict ourselves to what we call $\omega$-local (sente
and gote) sequences. 

We say that a move is $\omega$-local (with respect
to the previous move) if its gridcular distance from previous move
is smaller than a fixed number $\omega$; in this paper, we used $\omega=10$ for the strength
dataset and $\omega=5$ for the style dataset.
The simplifying assumption we make is that responses to
sente moves are always local. Although this does not hold in general, the
feature proves useful.

The assumption allows to partition each game into disjunct $\omega$-local sequences
(that is, each move in the
sequence is $\omega$-local with respect to its directly previous move) and observe
whether the player who started the sequence is different from the player who ended it.
If it is so, the $\omega$-local sequence is said to be sente for the player who started it
because he gets to play somewhere else first (tenuki). Similarly if the player who 
started the sequence had to respond last we say that the sequence is gote for him.
Based on this partitioning, we can count the average number of sente and gote
sequences per game from the sample $GC$ and these two numbers form the feature.

\subsection{Border Distance}
\label{feat:bdist}
The border distance feature is a two dimensional histogram counting the average number of moves
in the sample played low or high in different game stages. The original inspiration was
to help distinguishing between territorial and influence based moves in the opening, though
it turns out that the feature is useful in other phases of the game as well.

The first dimension is specified by the move's border distance,
the second one by the number of the current move from the beginning of the game.
The granularity of each dimension is given by intervals dividing the domains.
We use the
$$ByDist = \{ \langle1, 2\rangle, \langle 3 \rangle, \langle4\rangle, \langle 5, \infty)\}$$
division for the border distance dimension
(distinguishing between the first 2 lines, 3rd line of territory, 4th line of influence and
higher plays for the rest).
The move number division is given by
$$ByMoves_{STR} = \{ \langle1, 10\rangle, \langle 11, 64\rangle, \langle 65,200\rangle, \langle 201, \infty)\}$$ 
for the strength dataset and
$$ByMoves_{STYLE} = \{ \langle1, 16\rangle, \langle 17, 64\rangle, \langle 65,160\rangle, \langle 161, \infty)\}$$
for the style dataset. The motivation is to (very roughly) distinguish
between the opening, early middle game, middle game
and endgame. Differences in the interval sizes were found empirically and
our interpretation is that in the case of style,
we want to put bigger stress on opening and endgame (both of which can
be said to follow standard patterns)
on behalf of the middle game (where the situation is usually very complex).

If we use the $ByMoves$ and $ByDist$ intervals to divide the domains, we obtain a histogram
of total $|ByMoves| \times |ByDist| = 16$ fields. For each move in the games $GC$,
we increase the count in the
appropriate histogram field. In the end, the whole histogram is normalized
to establish invariancy under the number of games scanned by dividing the 
histogram elements by $|GC|$. The resulting 16 numbers form the border distance
feature.

\subsection{Captured Stones}
\label{feat:capt}
Apart from the border distance feature, we also maintain a two-dimensional histogram
which counts the numbers of captured stones in different game stages. The motivation is
simple -- especially beginners tend to capture stones because ``they could'' instead of
because it is the ``best move''. Such capture could
be a grave mistake in the opening and it would not be played by skilled players.

As before, one of the dimensions is given by the intervals 
$$ByMoves = \{ \langle1, 60\rangle, \langle 61, 240\rangle, \langle 241, \infty)\}$$
which try to specify the game stages (roughly: opening, middle game, endgame).
The division into game stages is coarser than for the previous
feature because captures occur relatively infrequently. Finer graining
would require more data.

The second dimension has a fixed size of three bins. Along the number of captives
of the player of interest (the first bin), we also count the number of his
opponent's captives (the second bin) and a difference between the two numbers
(the third bin). Together, we obtain a histogram of $|ByMoves| \times 3 = 9$ elements.

Again, the colored games $GC$ are processed move by move by increasing
the counts of captivated stones (or 0) in the appropriate field.
The 9 numbers (again normalized by dividing by $|GC|$) together comprise the feature.

\subsection{Win/Loss Statistic}
\label{feat:win}
The next feature is a statistic of wins and losses and whether they were
by points or by resignation.
The motivation is that many weak players continue playing games that are already lost
until the end, either because their counting is not very good (they do not
know there is no way to win), or because they hope the opponent will make a blunder.
On the other hand, professionals do not hesitate to resign if they think that
nothing can be done, continuing with a lost game could even be considered rude.

We disregard forfeited, unfinished or tie games in this feature
because the frequency of these events is so small it would
require a very large dataset to utilize them reliably.

In the colored games of $GC$, we count how many times did the player of interest:
\begin{itemize}
    \item win by counting,
    \item win by resignation,
    \item lost by counting,
    \item and lost by resignation.
\end{itemize}
Again, we divide these four numbers by $|GC|$ to maintain the invariance under the number of games
in $GC$. Furthermore, for the games won or lost in counting we count the average
size of the win or loss in points. The six numbers form the feature.

\section{Prediction}
\label{sec:mach}
So far, we have considered how we can turn a set of coloured games $GC$ into
an evaluation vector. Now, we are going to show how to utilize the evaluation.
To predict various player attributes, we start with a given input dataset $D$
consisting of pairs $D=\{ (GC_i, y_i), \dots\}$, where $GC_i$
corresponds to a set of colored games of $i$-th player and $y_i$ is the
target attribute. The $y_i$ might be fairly arbitrary, as long as it has
\emph{some} relation to the $GC_i$. For example, $y_i$ might be $i$'s strength.

Now, let us denote our evaluation process presented before as $eval$ and
let $ev_i$ be evaluation of $i$-th player, $ev_i = eval(GC_i)$. Then,
we can transform $D$ into $D_{ev} = \{(ev_i, y_i), \dots\}$, which forms
our training dataset.

As usual, the task of subsequent machine learning algorithm is to
generalize the knowledge from the dataset $D_{ev}$ to predict correct
$y_X$ even for previously unseen $GC_X$.  In the case of strength,
we might therefore be able to predict strength $y_X$ of an unknown
player $X$ given a set of his games $GC_X$ (from which we can compute
the evaluation $ev_X$).


\subsection{Prediction Model}
\label{sec:mach:pred}

Choosing the best performing predictor is often a tedious task, which 
depends on the nature of the dataset at hand, requires expert judgement
and repeated trial and error. In~\cite{Moudrik13}, we experimented with
various methods, out of which stacked ensembles~\cite{breiman96} with
different base learners turned out to have supreme performance.
Since this paper focuses on the evaluation rather than finding
the very best prediction model,
we decided to use a bagged artificial neural network,
because of its simplicity and the fact that it performs very well in
practice.  

The network is composed of simple computational units which are organized
in a layered topology, as described e.g. in monograph by~\cite{haykin_nn}.
We have used a simple feedforward neural network with 20~hidden units in one
hidden layer. The neurons have standard sigmoidal activation function and
the network is trained using the RPROP algorithm~\cite{Riedmiller1993} for at most
100~iterations (or until the error is smaller than 0.001). In both datasets 
used, the domain of the particular target variable (strength, style) was
linearly rescaled to $\langle -1,1 \rangle$ prior to learning. Similarly,
predicted outputs were rescaled back by the inverse mapping.

The bagging~\cite{breimanbag96} is a method that combines an ensemble
of $N$ models (trained on differently sampled data) to improve their
performance and robustness. In this work, we used a bag of $N=20$ above
specified neural networks.

\subsection{Reference Model and Performance Measures}
\label{sec:mach:ref}

In our experiments, mean regression was used as a reference
model. The mean regression is a simple method which constantly predicts the average of
the target attributes $\bar{y}$ in the dataset regardless of the particular
evaluation $ev_i$.
Although mean regression is a very trivial model,
it gives some useful insights about the distribution
of target variables $y$. For instance, low error of the mean
regression model raises suspicion that the target attribute $y$ is
ill-defined, as discussed in the results section of the 
style prediction, Section~\ref{res:style}.

%

To assess the efficiency of our method and give estimates of its precision for
unseen inputs, we measure the performance of our algorithm given a dataset $D_{ev}$.
A standard way to do this is to divide the $D_{ev}$ into training
and testing parts and compute the error of the method on the testing part.
For this, we have used a standard method of 10-fold cross-validation~\cite{crossval},
which randomly divides the dataset into 10
disjunct partitions of (almost) the same size.
Repeatedly, each partition is then taken as testing data,
while the remaining $9/10$ partitions are used to train the model.
Cross-validation is known to provide error estimates which are close to
the true error value of the given prediction model.

To estimate the variance of the errors, the whole 10-fold cross-validation process
was repeated 5 times, as the results in Tables~\ref{tab:str_reg_res}, \ref{tab:sty_feat_res} and \ref{tab:sty_reg_res} show.

A commonly used performance measure is the mean square error ($MSE$) which estimates variance of
the error distribution. We use its square root ($RMSE$) which is an estimate of
standard deviation of the predictions,
$$ RMSE = \sqrt{\frac{1}{|T_s|} \sum_{(ev, y) \in T_s}{ (predict(ev) - y)^2}} $$
where the machine learning model $predict$ is trained on the
training data $T_r$ and $T_s$ denotes the testing data.

\section{Experiments and Results}
\label{sec:expe}

\subsection{Strength}
One of the two major domains we have tested our framework on is the prediction of player
strength.

\subsubsection*{Dataset}
\begin{figure*}[htb]
\centering
\psfrag{ 10}[][b]{10}
\psfrag{ 15}[][b]{15}
\psfrag{ 20}[][b]{20}
\psfrag{ 25}[][b]{25}
\psfrag{ 30}[][b]{30}
\psfrag{ 35}[][b]{35}
\psfrag{ 40}[][b]{40}
\psfrag{ 45}[][b]{45}
\psfrag{ 50}[][b]{50}

\psfrag{-5.0}[][B]{}
\psfrag{-4.0}[][B]{5d}
\psfrag{-3.0}[][B]{}
\psfrag{-2.0}[][B]{3d}
\psfrag{-1.0}[][B]{}
\psfrag{0.0}[][B]{1d}
\psfrag{1.0}[][B]{}
\psfrag{2.0}[][B]{2k}
\psfrag{3.0}[][B]{}
\psfrag{4.0}[][B]{4k}
\psfrag{5.0}[][B]{}
\psfrag{6.0}[][B]{6k}
\psfrag{7.0}[][B]{}
\psfrag{8.0}[][B]{8k}
\psfrag{9.0}[][B]{}
\psfrag{10.0}[][B]{10k}
\psfrag{11.0}[][B]{}
\psfrag{12.0}[][B]{12k}
\psfrag{13.0}[][B]{}
\psfrag{14.0}[][B]{14k}
\psfrag{15.0}[][B]{}
\psfrag{16.0}[][B]{16k}
\psfrag{17.0}[][B]{}
\psfrag{18.0}[][B]{18k}
\psfrag{19.0}[][B]{}
\psfrag{20.0}[][B]{20k}

\includegraphics[width=\textwidth, height=150px]{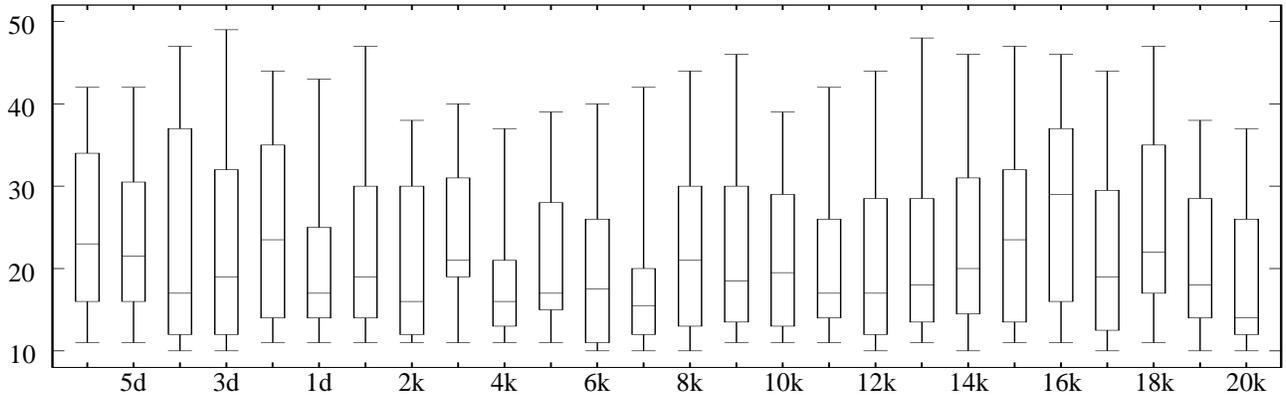}
\caption[Boxplot of game sample sizes]{Boxplot of game sample sizes. The box spans between 25th and 75th percentile,
the center line marks the mean value. The whiskers cover 95\% of the population.
The kyu and dan ranks are shortened to k and d.}
\label{fig:str_sizedist}
\end{figure*}

We have collected a large sample of games from the public archives of
the Kiseido Go server~\cite{KGSArchives}.  The sample consists of over
100 000 records of games in the \emph{.sgf} format~\cite{SGF}.

For each rank $r$ in the range of 6-dan to 20-kyu, we gathered a list of
players $P_r$ of the particular rank. To avoid biases caused by different
strategies, the sample only consists of games played on $19 \times 19$
board
between players of comparable strength
(excluding games with handicap stones).
The set of colored games $GC_p$ for
a~player $p \in P_r$ consists of the games player $p$ played when he
had the rank $r$. We only use the $GC_p$ if the number of games is not
smaller than 10 games; if the sample is larger than 50 games, we randomly
choose a subset of the sample (the size of subset is uniformly randomly
chosen from interval $\langle 10, 50\rangle$).  Note that by cutting
the number of games to a fixed number (say 50) for large samples, we
would create an artificial disproportion in sizes of $GC_p$, which could
introduce bias into the process. The distribution of sample sizes is
shown in Figure~\ref{fig:str_sizedist}.

For each of the 26 ranks, we gathered 120 such $GC_p$'s.  The target
variable $y$ to learn from directly corresponds to the ranks: $y=20$
for rank of 20-kyu, $y=1$ for 1-kyu, $y=0$ for 1-dan, $y=-5$ for 6-dan,
other values similarly. (With increasing strength, the $y$ decreases.)
Since the prediction model used (bagged neural network) rescales the
input data to $\langle -1,1 \rangle$, the direction of the ordering
or its scale can be chosen fairly arbitrarily.

\subsubsection*{Results}

The performance of the prediction of strength is given in
Table~\ref{tab:str_reg_res}. The table compares performances
of different features (predicted by the bagged neural network,
Section~\ref{sec:mach:pred}) with the reference model of
mean regression.

The results show that the prediction of strength has standard deviation
$\sigma$ (estimated by the $RMSE$ error) of approximately $2.66$ rank.
Comparing different features reveals that for the prediction of
strength, the Pattern feature works by far the best, while other
features bring smaller, yet nontrivial contribution.


\begin{table}
\begin{center}
\caption{
    RMSE errors of different features on the Strength dataset, a bagged neural
    network (Section~\ref{sec:mach:pred}) was used as the predictor in all but
    the first row.
    10-fold cross-validation was repeated 5 times to compute the errors
    and estimate the standard deviations.
    The last column shows multiplicative improvement in RMSE with regard to
    the mean regression.
} 
\label{tab:str_reg_res}
\begin{tabular}{|c|c|c|}
\hline
\textbf{Feature} & $\mathbf{RMSE}$ \textbf{error} & \textbf{Mean cmp}\\
\hline
None (Mean regression)  & $7.503 \pm 0.001$ & $1.00$\\
\hline
Patterns                & $2.788 \pm 0.013$ & $2.69$\\
$\omega$-local Seq.     & $5.765 \pm 0.003$ & $1.30$\\
Border Distance         & $5.818 \pm 0.010$ & $1.29$\\
Captured Stones         & $5.904 \pm 0.005$ & $1.27$\\
Win/Loss Statistic      & $6.792 \pm 0.006$ & $1.10$\\
Win/Loss Points         & $5.116 \pm 0.016$ & $1.47$\\
\hline
\textbf{All features combined} & $\mathbf{2.659 \pm 0.002}$ & $\mathbf{2.82}$\\
\hline

\end{tabular}
\end{center}
\end{table}

\subsection{Style}
\label{res:style}

The second domain is the prediction of different aspects of player styles.

\subsubsection*{Dataset}
\label{res:style:data}
The collection of games in this dataset comes from the Games of Go on
Disk database by~\cite{GoGoD}.  This database contains more than 70 000
professional games, spanning from the ancient times to the present.

We chose 25 popular professional players (mainly from the 20th century)
and asked several experts (professional and strong amateur players)
to evaluate these players using a questionnaire. The experts (Alexander
Dinerchtein 3-pro, Motoki Noguchi 7-dan, Vladim\'{i}r Dan\v{e}k 5-dan,
Luk\'{a}\v{s} Podp\v{e}ra 5-dan and V\'{i}t Brunner 4-dan) were asked to
assess the players on four scales, each ranging from 1 to 10.

\begin{table}[h!]
\begin{center}
\caption{The definition of the style scales.}
\label{tab:style_def}
\begin{tabular}{|c|c|c|}
\hline
\textbf{Style} & \textbf{1} & \textbf{10}\\ \hline
Territoriality & Moyo & Territory \\
Orthodoxity & Classic & Novel \\
Aggressivity& Calm & Fighting \\
Thickness & Safe & Shinogi \\ \hline
\end{tabular}
\end{center}
\end{table}

The scales (cf.~Table~\ref{tab:style_def}) try to reflect some of the traditionally perceived playing
styles.  For example, the first scale (\emph{territoriality}) stresses
whether a player prefers safe, yet inherently smaller territory (number 10
on the scale), or roughly sketched large territory (\emph{moyo}, 1 on the
scale), which is however insecure. For detailed analysis of playing
styles, please refer to~\cite{GoGoD:styles}, or~\cite{senseis:styles}.

For each of the selected professionals, we took 192 of his games from
the GoGoD database at random. We divided these games (at random) into 12
colored sets $GC$ of 16 games.  The target variable (for each of the four
styles) $y$ is given by average of the answers of the experts. Results
of the questionnaire are published online in~\cite{style_quest}. Please
observe, that the style dataset has both much smaller domain size and
data size (only 4800 games).

\subsubsection*{Results}
%

Table~\ref{tab:sty_feat_res} compares performances of different
features (as predicted by the bagged neural network, Section~\ref{sec:mach:pred})
with the mean regression learner. Results in the table have been averaged
over different styles. The table shows that the two features
with biggest contribution are the pattern feature and the border distance
feature. Other features perform either weakly, or even slightly worse than
the mean regression learner.
 
The prediction performance per style is shown Table~\ref{tab:sty_reg_res}
(computed on the full feature set). Given that the style scales have range
of 1 to 10, we consider the average standard deviation from correct answers
of around 1.6 to be a good precision.

We should note that the mean regression has very small $RMSE$ for
the scale of thickness.  This stems from the fact that the experts'
answers from the questionnaire have themselves very little variance. Our
conclusion is that the scale of thickness is not well defined. Refer
to~\cite{Moudrik13} for further discussion.

\begin{table}
\begin{center}
\caption{
    RMSE errors of different features on the Style dataset, a bagged neural
    network (Section~\ref{sec:mach:pred}) was used as the predictor in all but
    the first row.
    10-fold cross-validation was repeated 5 times to compute the errors
    and estimate the standard deviations.
    Both the RMSE scores and respective standard deviations were averaged over
    the styles.
    The last column shows multiplicative improvement in RMSE with regard to
    the mean regression.
} 
\label{tab:sty_feat_res}
\begin{tabular}{|c|c|c|}
\hline
\textbf{Feature} & $\mathbf{RMSE}$ \textbf{error} & \textbf{Mean cmp}\\
\hline
None (Mean regression)  & $2.168 \pm 0.002$ & $1.00$\\
\hline
Patterns                & $1.691 \pm 0.012$ & $1.28$\\
$\omega$-local Seq.     & $2.115 \pm 0.004$ & $1.03$\\
Border Distance         & $1.723 \pm 0.011$ & $1.26$\\
Captured Stones         & $2.223 \pm 0.025$ & $0.98$\\
Win/Loss Statistic      & $2.084 \pm 0.011$ & $1.04$\\
Win/Loss Points         & $2.168 \pm 0.005$ & $1.00$\\
\hline
\textbf{All features combined}   & $\mathbf{1.615} \pm 0.023$ & $\mathbf{1.34}$\\
\hline

\end{tabular}
\end{center}
\end{table}

\begin{table}[h]
\begin{center}
\caption{
    RMSE error for prediction of different styles using
    the whole feature set. A bagged neural network (Section~\ref{sec:mach:pred})
    was used as the predictor in all but the first row.
    10-fold cross-validation was repeated 5 times to compute the errors
    and estimate the standard deviations.
    The last column shows multiplicative improvement in RMSE with regard to
    the mean regression.
}
\label{tab:sty_reg_res}
\begin{tabular}{|c|c|c|c|}
\hline
& \multicolumn{2}{|c|}{$\mathbf{RMSE}$ \textbf{error}} & \\
\hline
\textbf{Style}  & \textbf{Mean regression} & \textbf{Bagged NN} & \textbf{Mean cmp} \\
\hline
Territoriality  & $2.396 \pm 0.003$ & $1.575 \pm 0.016$ & $1.52$ \\
Orthodoxity     & $2.423 \pm 0.002$ & $1.773 \pm 0.043$ & $1.37$ \\  
Aggressivity    & $2.183 \pm 0.002$ & $1.545 \pm 0.018$ & $1.41$ \\
Thickness       & $1.672 \pm 0.001$ & $1.567 \pm 0.015$ & $1.06$ \\

\hline
\end{tabular}
\end{center}
\end{table}

\section{Discussion}
\label{sec:disc}



In this paper, we have chosen the target variables to be the strength and
four different aspects of style. This has several motivations. Firstly,
the strength is arguably the most important player attribute and
the online Go servers allow to obtain reasonably precise data easily.
The playing styles have been chosen for their strong intuitive appeal to
players, and because they are understood pretty well in traditional Go
theory. Unlike the strength, the data for the style target variables are
however hard to obtain, since the concepts have not been traditionally
treated with numerical rigour. To overcome this obstacle, we used the
questionnaire, as discussed in Section~\ref{res:style:data}.

The choice of target variable can be quite arbitrary, as long as some
dependencies between the target variable and evaluations exist (and
can be learned). Some other possible choices might be the era of a player
(e.g. standard opening patterns have been evolving rapidly during the
last 100 years), or player nationality.

The possibility to predict player's attributes
demonstrated in this paper shows that the evaluations are a very useful
representation. Both the predictive power and the representation can
have a number of possible applications.


So far, we have utilized some of the findings in an online web
application\footnote{\url{http://gostyle.j2m.cz}}. It evaluates games submitted by players
and predicts their playing strength and style. The predicted strength
is then used to recommend relevant literature and the playing style is
utilized by recommending relevant professional players to review.
So far, the web application has served thousands of players and it was generally
very well received.
We are aware of only two tools, that do something
alike, both of them are however based on a predefined questionnaire. The first one
is the tool of~\cite{style:baduk} --- the user answers 15 questions and based 
on the answers he gets one of predefined recommendations.
The second tool is not available at
the time of writing, but the discussion at~\cite{senseis:which_pro}
suggests, that it computed distances to some pros based on user's
answers to 20 questions regarding the style. We believe that our approach
is more precise, because the evaluation takes into
account many different aspects of the games. 

Of course, our methods for style estimation are trained on very strong
players and thus they might not be fully generalizable to ordinary
players. Weak players might not have a consistent style, or the whole
concept of style might not be even applicable for them. Estimating this
effect is however not easily possible, since we do not have data about
weak players' styles.  Our web application allows the users to submit their own
opinion about their style, therefore we should be able to consider this
effect in the future research.

It is also possible to study dependencies between single elements
of the evaluation vector and the target variable $y$ directly. By
pinpointing e.g. the patterns of the strongest correlation with bad
strength (players who play them are weak), we can warn the users not
to play the moves associated with the pattern.
We have also realised this feature in the online web
application~\cite{GoStyleWeb}. However, this method seems to be usable
only for the few most strongly correlated attributes, the weakly correlated
attributes are prone to larger errors.

Other possible applications include helping the ranking algorithms to
converge faster --- usually, the ranking of a player is determined from
his opponents' ranking by looking at the numbers of wins and losses
(e.g. by computing an Elo rating~\cite{Elo}). Our methods might improve this by
including the domain knowledge.  Similarly, a computer Go program can
quickly classify the level of its human opponent based on the evaluation
from their previous games and auto-adjust its difficulty settings
accordingly to provide more even games for beginners. We will research
these options in the future.

\section{Conclusion}
\label{sec:conc}

This paper presents a method for evaluating players based on a sample
of their games. From the sample, we extract a number of different
domain-specific features, trying to capture different pieces of
information. Resulting summary evaluations turn out to be very
useful for prediction of different player attributes (such as
strength or playing style) with reasonable accuracy.

The ability to predict such player attributes has some very
interesting applications in both computer Go and in development
of teaching tools for human players, some of which we realized
in an on-line web application. The paper also discusses other potential
extensions and applications which we will be exploring in the future.

We believe that the applications of our findings can help
to improve both human and computer understanding of the game of Go.

\section{Implementation}
\label{sec:impl}

The code used in this work
is released online as a part of GoStyle project~\cite{GoStyleWeb}.
The majority of the source code is implemented in
the Python programming language~\cite{Python27}.

The machine learning models were implemented and evaluated using the
Orange Datamining suite~\cite{demsar13a} and
the Fast Artificial Neural Network library FANN~\cite{Nissen2003}.
We used the Pachi Go engine~\cite{Pachi} for the raw game processing.

\subsection*{Acknowledgment} 
This research has been partially supported by the Czech Science Foundation
project no.~P103-15-19877S. J. Moud\v{r}\'\i k has been supported by the
Charles University Grant Agency project no.~364015 and by SVV project
no.~260 224.

\section*{Glossary} 
\begin{itemize}
    \item \emph{atari} ---
        a situation where a stone (or group of stones) can be
        captured by the next opponent move,
    \item \emph{sente} ---
        a move that requires immediate enemy response, and thus keeps
        the initiative,
    \item \emph{gote} ---
        a move that does not require immediate enemy response, and thus loses
        the initiative,
    \item \emph{tenuki} ---
        a move gaining initiative -- ignoring last (gote) enemy move,
    \item \emph{handicap} ---
        a situation where a weaker player gets some stones placed on
        predefined positions on the board as an advantage to start the game with
        (their number is set to compensate for the difference in skill).
\end{itemize}

\bibliographystyle{IEEEtran}
\bibliography{clanek}

\end{document}